\def\year{2021}\relax
\documentclass[letterpaper]{article} 
\usepackage{aaai21}  
\usepackage{times}  
\usepackage{helvet} 
\usepackage{courier}  
\usepackage[hyphens]{url}  
\usepackage{graphicx} 
\usepackage{natbib}  
\usepackage{caption} 
\title{Learning Generalized Relational Heuristic Networks for Model-Agnostic Planning}
\author{

    Rushang Karia, Siddharth Srivastava
    \\
}
\affiliations{

    Arizona State University

}

\usepackage[dvipsnames]{xcolor}
\usepackage{import}
\usepackage{listings}
\usepackage{wrapfig}
\usepackage{multicol}
\usepackage{booktabs}
\usepackage{subcaption}
\usepackage{amsthm}
\usepackage{amsmath}
\newtheorem{definition}{Definition}[section]
\newtheorem{theorem}{Theorem}[section]
\usepackage{pgfplotstable}
\usepackage{graphicx}
\graphicspath{images/}

\usepackage{algorithm}
\usepackage[noend]{algpseudocode}

\usepackage{pdflscape}
\usepackage{afterpage}

\usepackage[utf8]{inputenc}
\usepackage{pgfplots}
\usepackage{tikzscale}
\DeclareUnicodeCharacter{2212}{−}
\usepgfplotslibrary{groupplots,dateplot}
\usetikzlibrary{patterns,shapes.arrows}
\pgfplotsset{compat=newest}

\usepackage{url}
\usepackage{amsthm}
\usepackage{amssymb}
\usepackage{enumitem}
\usepackage{lipsum}

\theoremstyle{definition}
\newtheorem{example}{Example}[section]

\usetikzlibrary{spy}
\usepackage{lineno}

\begin{document}

\maketitle

\begin{abstract}
Computing goal-directed behavior is essential to designing efficient AI systems. Due to the computational complexity of planning, current approaches rely primarily upon hand-coded symbolic action models and hand-coded heuristic-function generators for efficiency. Learned heuristics for such problems have been of limited utility as they are difficult to apply to problems with objects and object quantities that are significantly different from those in the training data. This paper develops a new approach for learning generalized heuristics in the absence of symbolic action models using deep neural networks that utilize an input predicate vocabulary but are agnostic to object names and quantities.  It uses an abstract state representation to facilitate data efficient, generalizable learning. Empirical evaluation on a range of benchmark domains show that in contrast to prior approaches, generalized heuristics computed by this method can be transferred easily to problems with different objects and with object quantities much larger than those in the training data. 
\end{abstract}
\section{Introduction}
\label{introduction}

The computational complexity of automated planning \cite{DBLP:conf/ijcai/Bylander91, DBLP:journals/ai/Bylander94} has motivated research on heuristics \cite{DBLP:journals/jair/HoffmannN01, DBLP:journals/ai/BonetG01, DBLP:conf/aips/HelmertD09}, that,
in conjunction with search algorithms can efficiently find a solution \cite{DBLP:journals/tssc/HartNR68, DBLP:journals/jair/HoffmannN01, DBLP:journals/ai/BonetG01}. The primary disadvantage of this technique is the initial investment required. Designing good heuristic-generation principles such as ``delete-relaxation'' \cite{DBLP:journals/jair/HoffmannN01} often requires a careful study of the representation language or the structure
of the underlying problems. These factors
make automatic synthesis of heuristics particularly attractive. 

There is evidence 
that one can learn such heuristics from the action models automatically when there is enough data \cite{DBLP:conf/aips/GroshevGTSA18, shen20:stripshgn}.
Existing approaches for learning such heuristics have proved to be difficult to scale and to transfer to problems with object names and object quantities not seen in the training data.
Moreover, such approaches often require hand-engineered action models in a representational
language such as the Planning Domain Definition Language (PDDL) \cite{DBLP:journals/jair/FoxL03} (see Sec. \ref{sec:related_work} for details on related work). Whether such synthesis can be done in the absence of action models or when sufficient training data is not present is a question that has not been sufficiently addressed.

This paper answers this question by presenting a method for learning generalizable heuristics when the domain model and/or 
domain expert are unavailable by utilizing a library of plans to train an artificial neural network.  We showcase the effectiveness of abstraction techniques as a domain-independent method for learning heuristic generation functions that do not require access to symbolic action models. We demonstrate the effectiveness of generalized heuristics learned using this method and show that they can transfer to problems with object quantities and/or names different that those in the training data. 

In the absence of sufficient training data, we develop and evaluate leapfrogging, a bootstrapping technique that was proposed in recent work \cite{DBLP:conf/aips/GroshevGTSA18} but has not been sufficiently developed and tested for learning generalized heuristics. We show that this technique is data-efficient and can be used to learn competitive generalized heuristics in the absence of externally generated training data.

This paper is organized as follows. Sec. \ref{sec:background} presents the necessary formal framework.
Sec. \ref{sec:lgh} defines the learning problem and describes our approach for learning followed by a description of using the learned heuristic for planning (Sec. \ref{sec:vanilla_training}).
Sec. \ref{sec:empirical_evaluation} discusses obtained results. Sec. \ref{sec:related_work} summarizes related work followed by conclusions (Sec. \ref{sec:conclusion}).

\section{Formal Framework}
\label{sec:background}

A planning \emph{problem} is a tuple $\Gamma = \langle O, P, A, s_{\emph{init}}, g, \delta \rangle$
where $O$ is a set of objects, $P$ is a set of predicates and $A$ is a set of unit-cost actions. For typed domains, we automatically compile in types as unary predicates.
The state space $S$ for a planning problem as defined above is the set of all possible assignments of truth values to predicates in $P$ instantiated with objects from $O$. $s_{\emph{init}} \in S$ is the initial state and $g$ is a goal condition expressed as a conjunctive first-order logic formula over the instantiated atoms. $\delta: S \times A \rightarrow S$ determines the transition function. Different planning problems from an application domain (e.g. Logistics) share the same $P$ and $A$ components and these components together define a planning \emph{domain}. While a number of representations have been developed to express domain-wide, ``lifted'' actions \cite{DBLP:conf/ijcai/FikesN71, DBLP:journals/jair/FoxL03, Sanner:RDDL, DBLP:conf/uai/SrivastavaRRC14}; such actions could also be implemented using arbitrary generative
models or simulators. We assume w.l.o.g., that an action $a \in A$ can be parameterized as $a(o_1, \ldots, o_n)$ where $o_1, \ldots, o_n \in O$; we do not place any representational requirements on the specifications of $A$.
A solution to $\Gamma$ is a plan $\pi = a_0, \ldots, a_{n-1}$ which is a sequence of actions inducing a trajectory $\tau = s_0, \ldots, s_n$ such that
$s_0 \equiv s_{\emph{init}}$, $\delta(s_i, a_i) = s_{i+1}$ and $s_n \models g$. The plan length $|\pi|_{s_i}$from a state $s_i$ is the number of states starting from $s_{i+1}$ in $\tau$. We will use $P^k$ to refer to the set of predicates with arity $k$ and $P^{k+}$ for those with arity $k$ or greater.

A planning \emph{heuristic} is a function $h : S \rightarrow \mathbb{R}^+_0 \cup \{\infty\}$, where $h(s)$  estimates the cost of reaching the goal state from a state $s$. The optimal heuristic $h^*(s)$ provides the optimal cost of
reaching the goal from $s$. Typically, search algorithms maintain a priority queue of promising paths and use the heuristic function to compute the keys in this queue \cite{DBLP:books/daglib/0023820}. For example, the utility value used in A* is $f(s) = g(s) + h(s)$ where $g(s)$ is the length of the path up to $s$ and $h(s)$ is the heuristic value of $s$; the node expanded is the one with the minimum value of $f(s)$ \cite{DBLP:journals/tssc/HartNR68}.

We use \emph{canonical abstractions} \cite{DBLP:journals/toplas/SagivRW02} for representing a concrete state such that information about object names and numbers is lifted by grouping them using abstraction predicates. Grouping together states can lead to certain predicates becoming imprecise. As a result, three-valued logic is used to represent truth values of predicates in an abstract state. We introduce canonical abstraction using the help of the following example.
\begin{example}
\label{example}
Consider the \emph{gripper} domain, which consists of two rooms and a robot equipped with a set of grippers to pickup or drop balls \cite{DBLP:journals/jair/LongF03}.\\Let $s_{\emph{eg}} = \{\emph{free}^1(g_1), \emph{at}^2(b_1, r_a), \emph{at}^2(b_2, r_b), \emph{robotAt}^1(r_a)\}$ be a state in a \emph{gripper} problem instance $\Gamma$ expressed in \emph{typed} PDDL with $O = \{ r_a, r_b, g_1, b_1, b_2\}$, and 
$g= \emph{at}^2(b_1, r_b) \land \emph{at}^2(b_2, r_b)$. Let $\emph{type}(\emph{gripper}) = \{ g_1 \}$, $\emph{type}(\emph{room}) = \{ r_a, r_b \}$ and $\emph{type}(\emph{ball}) = \{ b_1, b_2 \}$ be the  object types.
\end{example}

\theoremstyle{definition}
\begin{definition}{(Role)}
The role of an object $o \in O$ in a state $s$ is the set of unary predicates that it satisfies:
$\textit{role}(o) = \{ p^1|p^1 \in P^1, p^1(o) \in s\}$.
\end{definition}

For the state in Example \ref{example}, the role of the object $r_a$ is $\emph{role}(r_a) = \{ \emph{room}, \emph{robotAt}\}$ whereas 
$\emph{role}(r_b) = \{\emph{room}\}$. We will use 
$\psi(r) = \{ o|o \in O, \emph{role}(o) = r\}$ to denote the set of objects having a particular role $r$.
Thus, $\psi(\{ \emph{room} \}) = \{ r_b\}$, $\psi(\{ \emph{ball}\}) = \{ b_1, b_2 \}$, $\psi(\{\emph{room}, \emph{robotAt}\}) = \{r_a\}$ and
$\psi(\{\emph{gripper}, \emph{free}\}) = \{g_1\}$. The maximum number of possible roles in any state $s$ in any problem $\Gamma$ derived from a domain $D$ with $|P^1|$ unary predicates is $2^{|P^1|}$.

\theoremstyle{definition}
\begin{definition}{(Canonical Abstraction)}
\label{def:abstraction}
The canonical abstraction of a state $s = \{ p^k(o_1, ..., o_k)|p^k \in P, o_1, ..., o_k \in O\}$
is an abstract state $\overline{s} = \{ \overline{p}^k(\textit{role}(o_1), ..., \textit{role}(o_k))|\overline{p}^k \equiv p^k\}$.
Let $\mathcal{O} = \psi(\textit{role}(o_1)) \times \ldots \times \psi(\textit{role}(o_k))$ then $\overline{p}^k$ is defined
as follows:
\begin{itemize}
    \item $\overline{p}^k(\textit{role}(o_1), \ldots, \textit{role}(o_k)) = 0  \iff \forall (o_1, \ldots, o_k) \in \mathcal{O}$ $\quad p^k(o_1, \ldots, o_k) \notin s$.
    \item $\overline{p}^k(\textit{role}(o_1), \ldots, \textit{role}(o_k)) = 1  \iff \forall (o_1, \ldots, o_k) \in \mathcal{O}$ $\quad p^k(o_1, \ldots, o_k) \in s$.
    \item $\overline{p}^k(\textit{role}(o_1), \ldots, \textit{role}(o_k)) = \frac{1}{2}  \iff $
    \begin{align*} 
        & & &(\exists (o_1, \ldots, o_k) \in \mathcal{O} \quad p^k(o_1, \ldots, o_k) \in s) \, \land \\
        & & &(\exists (o_1, \ldots, o_k) \in \mathcal{O} \quad p^k(o_1, \ldots, o_k) \notin s).
    \end{align*}
\end{itemize}
\end{definition}

Let $r_0 = \{ \emph{gripper}, \emph{free}\}$, $r_1 = \{\emph{room}, \emph{robotAt}\}$, $r_2 = \{ \emph{room}\}$ and $r_3 = \{ \emph{ball} \}$ be the roles in the state $s_{\emph{eg}}$. The canonical abstraction of the state $s_{\emph{eg}}$ is the abstract state 
$\overline{s_{\emph{eg}}} = \{ \emph{free}^1(r_0), \emph{at}^2(r_3, r_1), \emph{at}^2(r_3, r_2), \emph{robotAt}^1(r_1)\}$. 
The truth values for predicates in $\overline{s_{\emph{eg}}}$ are $\emph{free}^1(r_0) = 1, \emph{at}^2(r_3, r_1) = \frac{1}{2},
\emph{at}^2(r_3, r_2) = \frac{1}{2}$ and $\emph{robotAt}^1(r_1) = 1$.

This formulation assumes that domains contain unary and binary predicates. Domains with ternary or higher arity predicates can be easily compiled into domains with binary predicates. The framework presented in this paper can handle higher arity predicates, however, we found that the results were best in domains compiled as binary predicates. We present a case study of Sokoban in the supplementary material by compiling ternary predicates present in the Sokoban domain into binary predicates; Sokoban2. Learning such features is an independent problem and an active area of research \cite{DBLP:conf/aaai/BonetFG19}. These approaches could be used to learn such predicates for better abstractions. 

\section{The Generalized Heuristic Learning Problem}
\label{sec:lgh}

\begin{figure*}[t]
\centering
\includegraphics[width=\textwidth]{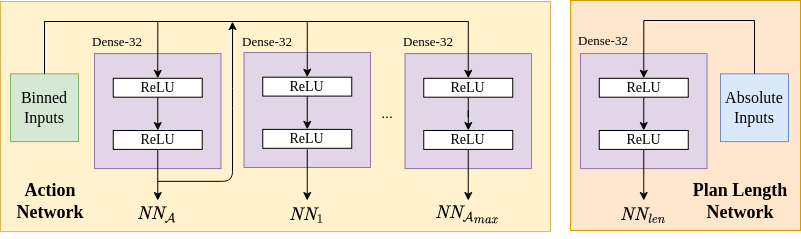}

\caption{ The network architecture used in this paper. Activations for $\emph{NN}_\mathcal{A}, (\emph{NN}_1, \ldots, \emph{NN}_{\mathcal{A}_{\emph{max}}})$ and $\emph{NN}_{\emph{len}}$ are SoftMax, Sigmoid and ReLU respectively. Each \emph{Dense-32} block contains two fully-connected hidden layers with 32 tensors, each using ReLU as the activation function. Absolute and Binned Inputs comprise vectors $v, {m}_{p_1}^2, \ldots, {m}_{p_n}^2$ and $v', {m'}_{p_1}^2, \ldots, {m'}_{p_n}^2$ respectively (described in Sec. \ref{sec:network_arch}).}
\label{fig:network}

\end{figure*}

We define the learning problem as follows:
\begin{definition}{(Learning Generalized Heuristics)}
Given a library of trajectories of the form $\Xi = \{ \langle\pi, \tau, g, O\rangle \}$ for a domain $D = \langle P, A\rangle$
where $O$ is a set of objects, $g$ is a goal formula, $\tau=s_0, \ldots, s_n$, $\pi = a_0, \ldots, a_{n-1}$ contain states and parameterized actions from a planning
problem $\langle D, O, s_{\textit{init}}, g, \delta \rangle$ such that $s_0 \equiv s_{\textit{init}}, \delta(s_i, a_i) = s_{i+1}$ and $s_n \models g$, learn a
domain-wide generalized heuristic function $h_D$ s.t. $h_D(s, g', O')$ estimates, for any planning
problem $\Gamma' = \langle D, O', s_{\textit{init}}', g', \delta'\rangle$ and any state $s$ in the state space of  $\Gamma'$, the distance
from $s$ to a state $s'$ s.t. $s' \models g'$.
\end{definition}

Our overall
approach for model-agnostic planning involves solving the learning problem defined above by training a Generalized Heuristic Network (GHN) (Sec. \ref{sec:lgh}) and using the learned GHN for planning (Sec. \ref{sec:vanilla_training}). Our approach is a domain-independent method for learning heuristic generation functions (HGFs) using either training data or problem generators. In the standard planning paradigm, this approach plays a role similar to that of HGFs, which are currently hand-coded. The computation of our heuristic is model-agnostic in that it only needs access to the action names and parameters, true atoms of a state, and the objects of the problem, which could be provided by a blackbox simulator.

\emph{Vanilla training data generation} To gather the training data $T$,
we first generate a set of problem instances and use an off-the-shelf solver to compute a plan
for each problem to form a library of trajectories $\Xi = \{\langle\pi, \tau, g, O\rangle\}$.  Next, for each trajectory $\xi \in \Xi$, we encode goal hints to every state $s \in \tau_{\xi}$ using the approach in Sec. \ref{sub:encode_aux} to form tuples $(s, a, |\pi|_s)$ which are then converted to
$(s, \overline{s}, a, |\pi|_s)$ using canonical abstraction (Definition \ref{def:abstraction}) and added to $T$. As a part of the data generation process, we maintain a set of roles $\mathcal{R}$, actions $\mathcal{A}$, the maximum number of action parameters $\mathcal{A}_{\emph{max}}$, and predicates $\mathcal{P}$ that occurred in the training data. Together, they
define the input-output dimensions of the network.
Once $T$ has been generated, we use standard optimization techniques to minimize the loss. 

\emph{Training data generation using leapfrogging} The training data generation method discussed above assumes access to a planner that can solve the problems in the training set.
It might be the case that access to such an oracle is either expensive or unavailable. If a problem generator is available, leapfrogging \cite{DBLP:conf/aips/GroshevGTSA18} can be used to incrementally generate training data in an iterative fashion. Initially, problem instances with very few objects $\Gamma'_0$ are solved to generate training data $T_0$. These instances are small enough that blind search can be used to find solutions. We then apply our learning approach to learn a GHN $\emph{leap}_0$ using $T_0$. Next, $\emph{leap}_0$ is used to bootstrap the generation of training data $T_i$ for the next iteration. We use the problem generator to generate problem instances of sizes $\Gamma'_{0}, \ldots, \Gamma'_{i}$ where problems in $\Gamma'_{i}$ have more objects than those in $\Gamma'_{i-1}$ and generate $T_{i}$ by using $\emph{leap}_{i-1}$ to 
solve $\Gamma'_{0}, \ldots, \Gamma'_{i}$. We then learn a new GHN $\emph{leap}_{i}$ using $T_{i}$. Since GHNs learn knowledge independent of the number of objects, this iterative approach allows GHNs to effectively scale even in the absence of efficient, off-the-shelf mechanisms for generating the seed plans.

\subsection{Network Architecture}
\label{sec:network_arch}

The neural network used in this paper is illustrated in Fig. \ref{fig:network}. We use two networks; one to predict the action and its parameters and the other to predict the plan length. We found this approach to be the best in providing good estimates of both the action probabilities and the plan length. We refer the reader to supplementary material for our ablation study.

The output of the network is a vector $\emph{NN}_\mathcal{A}$ of length $|\mathcal{A}|$ representing the action probability, a set of vectors $\emph{NN}_{1}, ..., \emph{NN}_{\mathcal{A}_{\emph{max}}}$ each of length $|\mathcal{P}^1|$ that
represents the predicted role of the corresponding parameter in the action (recall that a role is a set of unary predicates) and a real-valued number $\emph{NN}_{\emph{len}}$ that represents the predicted plan length.

The input to the neural network is an abstract state that is represented as a set of vectors and matrices which capture the abstraction of object properties as well as their relationships. We compute inputs of two different types: (a) Absolute Inputs and (b) Binned Inputs. Absolute inputs encode the actual counts of
the roles in a concrete state and also capture the role count of the $k$-ary atoms that are true in the state. For a concrete state $s$ and the corresponding abstract state $\overline{s}$ we represent all roles occurring in $s$ as a vector $\upsilon$ of length $|\mathcal{R}|$. Each $k$-ary predicate $p^k \in \mathcal{P}^{2+}$ is encoded as a matrix $m^k_p$ of dimensions $|\mathcal{R}|^{k} = |\mathcal{R}|_1 \times \ldots \times |\mathcal{R}|_k$. To encode absolute inputs, (a) $v[r]$ is set to the role count $|\psi(r)|$ for every role $r \in \mathcal{R}$ and, (b) $m_p^k[r_i, \ldots, r_j]$ is set to the number of tuples in $\psi(r_i) \times \dots \times \psi(r_j)$ such that $p(o_i, \ldots, o_j)$ is true in $s$.

Absolute inputs help in predicting the plan length since they capture information about the number
of objects in a role. However, for predicting actions, this low level
of granularity is unnecessary and we found that this can lead to poor accuracy in predicting the actions. Instead, we compute binned inputs $\upsilon'$ and ${m'}^k_p$ by categorizing the absolute inputs $\upsilon$ and $m^k_p$ into \emph{levels} -- which is a configurable hyperparameter -- that can express information about the structure
of the state at a higher level of granularity.  To encode binned
inputs, in our experiments, we (a) encoded $v'[i]$ as $min(v[i], 2)$ to categorize $\psi(r)$ as containing zero, one or more than one objects and, (b) encoded ${m'}^k_p[r_i, \dots, r_j]$ as one of the three truth values of the predicate $\overline{p}^k(r_i, \ldots, r_j)$ (as defined in Definition \ref{def:abstraction}) in $\overline{s}$. We experimented with a larger set of discrete values to categorize the absolute inputs, like $min(v[r], n \in \mathbb{N})$, which yielded
no significant improvement in network predictions for our experiments.

\subsection{Encoding Goal Hints}
\label{sub:encode_aux}

Inclusion of goal-relevant information has been shown to facilitate
learning goal dependent concepts \cite{DBLP:conf/icml/WinnerV03, DBLP:conf/aips/GroshevGTSA18}. We propose a simple scheme to encode goal hints for canonical abstractions that involves post-processing a state to add new unary predicates, without using domain knowledge, and whose complexity is linear in the number of atoms in the state and goal.

For the state $s_{\emph{eg}}$ in Example \ref{example} where $g = \emph{at}^2(b_1, r_b) \land \emph{at}^2(b_2, r_b)$ we add atoms $\emph{goal}_{\emph{at}}^2(b_1, r_b)$ and $\emph{goal}_{\emph{at}}^2(b_2, r_b)$ to $s_{\emph{eg}}$. This allows the network to identify \emph{goal} predicates. Since ${\emph{at}}^2(b_2, r_b) \in s_{\emph{eg}}$ we also add $\emph{done}_{\emph{at}}^2(b_2, r_b)$ to $s_{\emph{eg}}$ which further allows the
network to better identify relational structures of a state. For $\emph{at}^2(b_1, r_b)$ we add two unary atoms $\emph{goal}_{\emph{at}_1}^1(b_1)$ and $\emph{goal}_{\emph{at}_2}^1(r_b)$. We similarly add two other unary atoms for $\emph{at}^2(b_2, r_b)$. Doing so changes $\emph{role}(r_b)$ from $\{ \emph{room} \}$ to $\{\emph{room}, \emph{goal}_{{\emph{at}}_2} \}$ and $\emph{role}(b_1)$ from $\{ \emph{ball} \}$ to $\{\emph{ball}, \emph{goal}_{{\emph{at}}_1}\}$. These changes in object roles allow a richer representation of the abstract state since new roles demarcating objects which are part of goals have been introduced. Finally, since ${\emph{at}}^2(b_2, r_b) \in s_{\emph{eg}}$ and there is no other atom $\emph{at}$ appearing in the goal where $b_2$ is the first parameter, $\emph{done}^1_{\emph{at}_1}(b_2)$ is added to $s_{\emph{eg}}$ indicating that all atoms named $\emph{at}$ in $g$  where $b_2$ appears as the first parameter are satisfied in the current state.

In general, let $G$ refer to atoms in $g$ for a problem $\Gamma$. For an atom $p^k(o_1, \ldots, o_k) \in G$ add a new atom $\emph{goal}_{p}^k(o_1, \ldots, o_k)$ to the state. 
This captures goal related relational information in the state $s$. Also add a set of atoms $\cup_{i=1}^{k} \{\emph{goal}_{p_i}^1(o_i)\}$ to the state. As a consequence, an object appearing appearing only in $G^{2+}$ now gets a defined role in $s$. Whenever a goal atom $p^k(o_1, \ldots, o_k) \in s$, $\emph{done}_p^k(o_1, \ldots, o_k)$ is
added to the state, else it is removed. Additionally, for an object $o$, when 
$\exists p^k \in G \text{, } \exists i \in \mathbb{N}_1^k  \text{, }  \forall p^k(\ldots, o_i, \ldots) \in G \text{ } p^k(\ldots, o_i, \ldots) \in s$ where $o_i = o$ is satisfied, $\emph{done}_{p_i}^1(o)$ is added to the state, else it is removed. Intuitively, this means that an object at index $i$ for a predicate $p$ is marked as done iff all goal atoms named $p$ where the object appears at index $i$ are satisfied in the concrete state.

\section{Planning Using Generalized Heuristic Networks}
\label{sec:vanilla_training}

\emph{Hybrid heuristic function} Using just the action probability in an algorithm like policy-rollout \cite{DBLP:conf/ijcai/YoonFG07} can lead to poor performance since it does not provide estimates on the cost to reach the goal. Generally, the neighborhood of a state has low variance in terms of the plan length predicted since (a) different actions can lead to states encapsulated by the same abstract state, and (b) different abstract states in the neighborhood are not substantially different. This can cause plan length based search to get stuck expanding nodes in local minima. We mitigate these limitations by combining them to form a hybrid heuristic.

We define the \emph{artificial} path cost $g'(s)$ to be the sum of the action probabilities along the path to $s$ which we use to increase the path cost of low confidence paths. This, in conjunction with the predicted plan length helps arrive at better estimates concerning the nodes to expand. Since path information is typically stored in nodes, we compute $g'(\emph{node})$ and $h(\emph{node})$ from the network output as follows:

\begin{align}
V_{o+}(i, o) &= \frac{\sum\limits_{u_j \in \mathcal{P}^1 \cap \emph{role}(o)} f(\emph{NN}_{i}[u_j], \epsilon)}{|\mathcal{P}^1|} \\
V_{o-}(i, o) &= \frac{\sum\limits_{u_j \in \mathcal{P}^1 \setminus \emph{role}(o)} f(1 - \emph{NN}_{i}[u_j], \epsilon)}{|\mathcal{P}^1|} \\
V_p(i, o) &= V_{o+}(i, o) + V_{o-}(i, o) \\
V_a(a(o_1, ..., o_n)) &= 1 - \emph{NN}_\mathcal{A}[a] \times \frac{\sum_{i=1}^{n} V_p(i, o_i)}{n}\\
g'(\emph{node}) &= g'(\emph{node.parent}) + V_a(\emph{node.action}) \\
h(\emph{node}) &= \emph{NN}_{\emph{len}}
\end{align}

where $\emph{role}(o)$ is the role of the object $o$ in $\emph{node.state}$, $\epsilon \in [0, 1]$ is a threshold and $f$ is a filter: $\emph{f}(x, \epsilon) = 1$ if $ x \geq \epsilon$ and 0 otherwise. $V_a \in [0, 1]$ is the score of an action computed by determining the probability of the action along with the confidence of the instantiated parameters.  $V_{o+}(i, o)$ and $V_{o-}(i, o)$ compute the score of the parameterized object's role relative to the predicted role. The score of the instantiated parameter $o_i$, $V_p \in [0, 1]$ is a ratio of the total number of unary predicates that were correctly predicted for $\emph{role}(o_i)$. A low value of the action score indicates that the path is detrimental to search and should be penalized.

\emph{Searching using the learned heuristic network} GHNs can be used in standard graph-search based search algorithms like A* using a blackbox simulator for action application and retrieving the atoms of a state.  Given a node in the A* search tree, we use the hybrid heuristic, $f_\emph{GHN}(\emph{node}) = g'(\emph{node}) + h(\emph{node})$ described above, to determine which node to expand next.

Using $f_\emph{GHN}$ as the key in the priority queue in a search algorithm like A* only changes the order in which A* expands nodes. The actual (or real) path cost, $g(\emph{node.state})$ is used to determine
if a visited state has been reached by a cheaper path under standard operation of A*.
The following result follows from the
properties of A* when used with a closed list \cite{DBLP:books/daglib/0023820}.

\begin{theorem}
Planning with A* using $f_\textit{GHN}$ is sound and complete on finite state spaces.
\end{theorem}
\section{Empirical Evaluation}
\label{sec:empirical_evaluation}

We conducted an extensive evaluation using benchmarks from the International Planning Competition (IPC) \cite{DBLP:journals/jair/LongF03}. Our setup is similar to the closest published work \cite{shen20:stripshgn} in the area and contains a much larger suite of test problems. The source code along with the data reported is available in the supplementary material.

To the best of our knowledge, there is no source code available for HGFs that can be used without symbolic action models or other learning related work \cite{shen20:stripshgn, arxiv:GPDRL}. Hence, we compare our approach with planners that use hand-coded action models and hand-coded HGFs for computing heuristics. We implemented our approach within Pyperplan \cite{alkhazraji-et-al-zenodo2020} which is written in Python and is a common platform for implementing and evaluating algorithms. Most IPC winners however, use an optimized implementation in C/C++. We also evaluate our approach against IPC winners and show that our approach remains competitive.

Our results indicate that even though they do not use action models (a) GHNs are competitive when compared against hand-coded HGFs, (b) GHNs successfully transfer to problems with more objects than those in the training data, and (c) In the absence of externally generated training data, leapfrogging is an effective bootstrapping technique for data-efficient learning. We discuss the configuration and methods used for evaluating these hypotheses below.

\subsection{Empiricial Setup}

Our hardware configuration consists of a cluster of Intel Xeon E5-2680 v4 CPUs with 28 CPU's. We utilized all cores, however each problem was solved on a single core.

\begin{figure*}[t!]
\centering
\includegraphics[width=\textwidth]{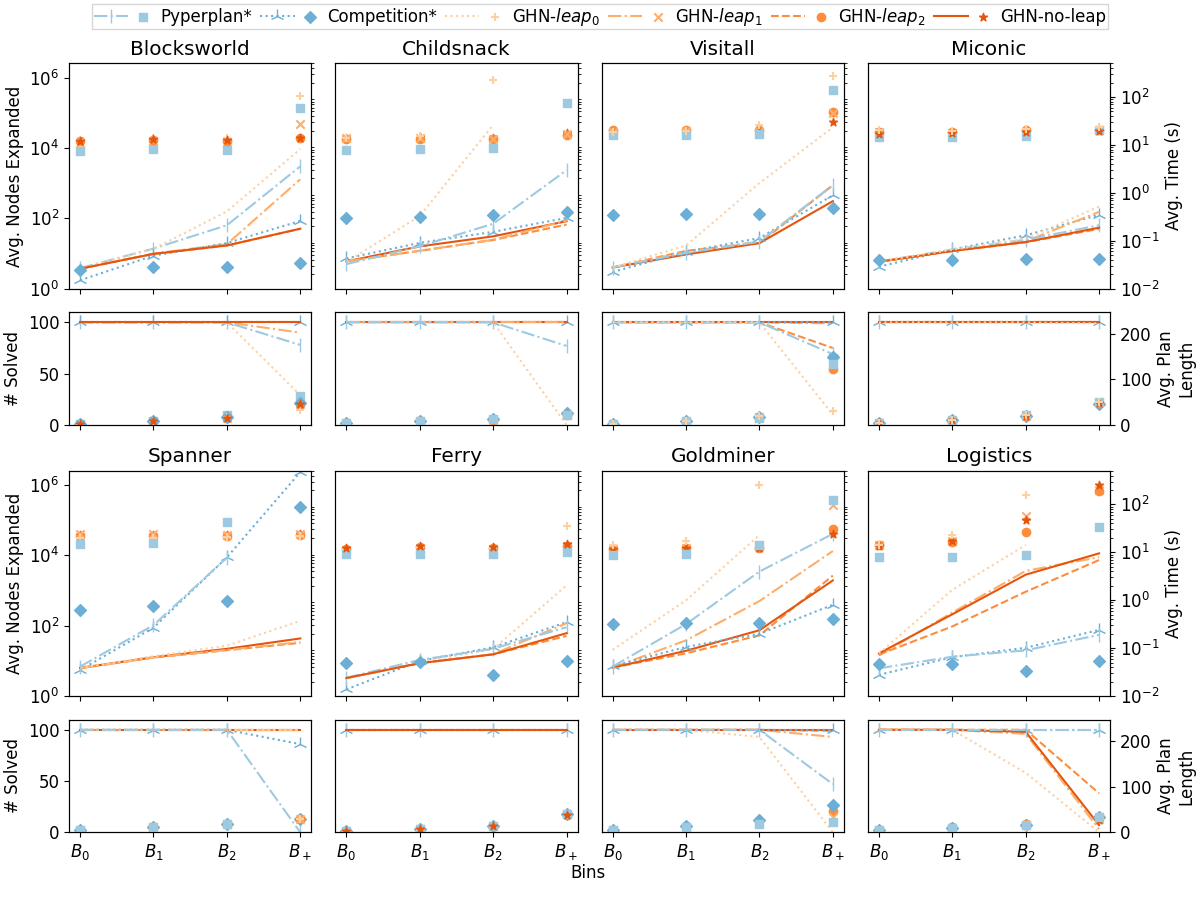}
\caption{Empirical results for our experiments. The x-axis represents the bins for that domain. Parameters for each bin can be found in the supplementary material. For the top plot, the line plot represents the avg. nodes expanded while the scatter plot represents the avg. time taken (in seconds) to solve a problem. For the bottom plot, the line plot represents the total number of problems solved (100 per bin) within the time limit while the scatter plot represents the avg. plan length.}
\label{fig:results}
\end{figure*}

Our evaluation consists of 13 benchmark domains from the IPC: \emph{Blocksworld, Childsnack, Ferry, Goldminer, Grid$\clubsuit$, Gripper$\clubsuit$, Grippers$\clubsuit$, Logistics, Miconic, Sokoban$\clubsuit$, Sokoban2$\clubsuit$, Spanner,} and \emph{Visitall}. We compare our approach with 6 baselines \{A*, GBFS\} + \{hff, lmcut\}, FF, and FD LAMA. \{A*, GBFS\} + \{hff, lmcut\} are implementations of the \emph{hff} \cite{DBLP:journals/jair/HoffmannN01} and \emph{lmcut} \cite{DBLP:conf/aips/HelmertD09} heuristics in Pyperplan using A* and Greedy Best First Search (GBFS) as the search algorithm. FF is a well-known competition winner implemented in C \cite{DBLP:journals/jair/HoffmannN01}. FD LAMA, the lama-first \cite{DBLP:journals/jair/RichterW10} configuration of Fast Downward \cite{DBLP:journals/jair/Helmert06}, is a state-of-art competition planner written in C++. Our approach is labeled as GHN-$\emph{leap}_i$.

The focus of this work is satisficing planning so our evaluation metrics focus on the number of nodes expanded and problems solved. We generated 400 test problems and the time limit was set to 600 seconds per problem. There were no restrictions on memory usage.
Since no single baseline outperforms the others in every domain, we compare our approach against the best performing Pyperplan implementation, labeled Pyperplan*, and competition planner, labeled Competition*, for each domain. For each domain, we used the same search algorithm as Pyperplan* during training and evaluation for GHNs. 

Our results are shown in Fig \ref{fig:results}. GHNs outperform Pyperplan*(Competition*) in 5(1) of 13 domains, are competitive on 5(8) domains and underperformed on 3(4). Representatives of all of these categories are included in the analysis here. Due to space constraints, analysis for domains labeled $\clubsuit$ above are included in the supplementary material.

We generated problems randomly from problem generators used by the organizers \cite{source:generator}. Problem sizes were scaled by increasing the number of objects along multiple dimensions in the generator parameters. This does not necessarily increase the difficulty but does increase the size of the state space. 

We categorized these sets of increasing problem sizes into ``bins" to showcase how leapfrogging can learn heuristics with just a problem generator. The bins were
indexed as $B_0, B_1$, and $B_2$ with number of objects monotonically increasing. The $i^{th}$ leapfrog iteration, GHN-$\emph{leap}_i$, was trained on problem sizes ranging in $B_0, \ldots, B_i$ using GHN-$\emph{leap}_{i-1}$ to generate the training plans. Training data for GHN-$\emph{leap}_0$ was generated using FF but even a blind solver could be used. GHN-$\emph{no-leap}$ is a special non-leapfrog network which was trained on problems in $B_0, \ldots, B_2$ by using FF to generate the training data. To showcase transfer learning, we use $B_{+}$ to denote problems containing number of objects that was greater than that in the training data. For example, in the Visitall domain we divided the problems based on the size $n$ of the square grid $b$; $B_0$: $n = 2$, $B_1$: $n = 3$, $B_2$: $n = 4$, $B_{+}$: $n \in \{5, \ldots, 15\}$. Our test set contains 100 problems per bin.

We used the architecture illustrated in Fig. \ref{fig:network} to train domain-wide GHNs. Our optimization algorithm was the Keras \cite{chollet2015keras} implementation of RMSProp \cite{RMSProp} configured with a learning rate, $\eta=0.001$ and $\hat{\epsilon}=1e-3$. GHNs were trained for 100 epochs with a batch size of 32. \emph{categorical cross entropy, binary cross entropy} and \emph{mean absolute error} were the loss minimization functions for the  $\emph{NN}_\mathcal{A}$, $\emph{NN}_{1, \ldots, \mathcal{A}_{\emph{max}}}$, and $\emph{NN}_{\emph{len}}$ layers respectively. Training data for GHN-$\emph{leap}_0$, GHN-$\emph{leap}_1$, GHN-$\emph{leap}_2$ and GHN-$\emph{no-leap}$ consisted of 100, 200, 400 and 400 problems respectively.

\subsection{Analysis}

\emph{(a) GHNs are competitive when compared against hand-coded HGFs} It is clear from Fig. \ref{fig:results} that despite not having access to symbolic action models and hand-coded HGFs, GHNs are comparable against approaches using action models and hand-coded HGFs. GHNs outperform Pyperplan* on 5 of the 13 domains presented, both in the number of nodes expanded and problems solved while producing plans of similar quality and are competitive on 5 domains. The number of nodes expanded by GHNs are magnitudes less than that of Pyperplan*. This difference is small enough in bins $B_0, \ldots, B_2$ that the avg. time to solve a problem is slightly higher for GHNs due to overheads like loading the network. However, the advantages of GHNs become apparent in $B_+$ where GHNs can solve more problems, often requiring less time per problem despite using neural network inference to compute the heuristic value.

GHNs are also competitive when compared against Competition* in 8 of the 13 domains, often expanding similar number of nodes while yielding the same plan length and solving the same number of problems. Competition planners outperform GHNs in the time taken to solve a problem, often solving problems instantaneously. A notable exception is the Spanner domain, where Competition* was unable to solve many problems in $B_+$ and required more time to solve the problems than GHNs. The Spanner domain was specifically designed to not work well with ``delete-relaxation" heuristics like those used in Competition*. This indicates that GHNs are able to learn knowledge of the problem structure that is orthogonal to existing heuristic generation concepts.

\emph{(b) GHNs successfully transfer to problems with more objects than those in the training data} As can be seen in  Fig. \ref{fig:results}, even though GHNs do not have access to action models, GHN-$\emph{leap}_2$ and GHN-$\emph{no-leap}$ easily transfer to problems in $B_+$ which consist of a larger number of objects than those in the training data. This highlights the advantages of abstraction techniques which can be used to learn HGFs that easily transfer to problems with more objects, and can be used even in the absence of action models.

\emph{(c) In the absence of externally generated training data, leapfrogging is an effective bootstrapping technique for data-efficient learning} Our results show that leapfrogging is data-efficient and can learn heuristics that are comparable to GHN-$\emph{no-leap}$ which used externally generated training data. We analyze leapfrogging by considering the Goldminer domain where GHN-$\emph{no-leap}$, whose training data was generated by using FF significantly outperforms Pyperplan* in $B_+$. GHN-$\emph{leap}_0$ which was the first iteration of leapfrogging was unable to outperform Pyperplan* on the test set since its ability to generalize was limited. However, the next iteration, GHN-$\emph{leap}_1$ whose learning process was bootstrapped by using GHN-$\emph{leap}_0$ to generate plans for GHN-$\emph{leap}_1$ outperforms Pyperplan*. The final leapfrog iteration, GHN-$\emph{leap}_2$ converges performance with GHN-$\emph{no-leap}$ despite using minimal external training data. This shows that leapfrogging is an effective domain-independent bootstrapping technique for learning generalized heuristics.

GHNs appear to perform best in domains whose problems have structured solutions. We now discuss results on select domains where GHNs did not outperform the baselines. GHNs could not generalize well on the Logistics domain and were outperformed by every baseline. We investigated the reasons for the poor performance and found that one of the reasons was the nature of training data produced. The plans for Logistics are quite diverse leading to a large network loss and consequently poor search performance. One reason for this diversity could be due to the tighter coupling of objects in Logistics as is mentioned in Rivlin et. al \cite{arxiv:GPDRL}.

While GHNs outperform Pyperplan*, they were unable to compete with Competition* on the Goldminer domain. One of the reasons is that GHNs do not capture landmarks very well. Goldminer has a very simple strategy where one needs to reach the correct $y$ location in a grid with the right tools, and then simply move their $x$ location to reach the goal. Landmarks can be used to solve problems in this domain relatively easily. This information is missing in our goal encoding scheme and could be used to improve performance by learning landmarks as well. Another way to mitigate this would be to use first-order logic with transitive closure FO(TC) when encoding the relations so that such ``location" related goal information can be captured in states that are ``far-away" but logically related. 

We observed that GHNs have a higher network loss when actions only change binary predicates. These actions only affect the relational inputs and not the vector role counts. As a result, predictions usually have a larger error which can become quite sensitive when the number of objects is small. For example, for problems in the Spanner domain with 8 spanners but only 1 nut to tighten, GHNs had a larger test error for the predicted plan length and hence expanded more nodes. However, as the number of objects increase, the test loss reduces enabling GHNs to outperform all other baselines including Competition*.

Our results show that in a similar search setting and once the problem state spaces grow large enough, GHNs outperform Pyperplan* in the time required to solve a problem. While the computational costs of heuristic estimates using hand-coded HGFs for these problems remains fixed, the computational cost of GHNs has plenty of room for improvements. One such improvement in our implementation would be to eliminate the data structure conversion overhead that was added as a result of using FastDownward's PDDL parser for our internal state representation. Other optimizations such as reducing network inference costs will naturally reduce the time required to solve a problem and will bridge the gap in differences with Competition*.

\section{Related Work}
\label{sec:related_work}

Our work builds upon the broad literature on learning for planning \cite{DBLP:journals/ker/JimenezRFFB12, DBLP:journals/ker/CelorrioAJ19}. Our approach relates the most closely with other methods for learning for planning that utilize deep learning.

Value iteration networks \cite{DBLP:conf/nips/TamarLAWT16} embed the standard value iteration computation within the network. While this method demonstrates successful learning, it encodes the input as an image limiting its effectiveness in solving problems whose states do not have a natural representation as images. Groshev et. al \cite{DBLP:conf/aaaiss/GroshevTGSA18} learn generalized reactive policies and heuristics using a
convolutional neural network (CNN). One drawback of their approach is that their network architecture and input feature vector representation are domain dependent and require a domain expert to provide them. ASNets \cite{DBLP:conf/aaai/ToyerTTX18} learn generalized policies by a network composed of alternating action and proposition layers. ASNets have a fixed receptive field that can potentially limit generalizability. STRIPS-HGNs \cite{shen20:stripshgn} learn domain-independent HGFs
by approximating the shortest path over the delete-relaxed hypergraph of a STRIPS \cite{DBLP:conf/ijcai/FikesN71} problem. To do this, 
they define a Hypergraph Network Block, utilizing message passing to increase the receptive field of the network. The generalizability of their network depends on the number of
message passing steps which can be a limiting factor as problem sizes scale up to much larger than the training data.  GBFS-GNNs \cite{arxiv:GPDRL} learn policies using network blocks similar to STRIPS-HGNs  but do not use the delete-relaxed version of the problem. Since they do not learn heuristics, they use rollout during search. A common limitation of ASNets, STRIPS-HGNs and GBFS-GNNs is that they require access to symbolic action models expressed in a language such as PDDL \cite{DBLP:journals/jair/FoxL03}. 

Curriculum learning \cite{DBLP:conf/icml/BengioLCW09} shows that effective learning is possible by organizing the training data in the form of a schedule. However unlike leapfrogging, this method assumes that training data is available. Bootstrap learning \cite{DBLP:conf/socs/ArfaeeZH10} incrementally learns a heuristic for solving a class of problems using the previous heuristic to generate training data for the next iteration. However, the learned heuristic cannot generalize to problems with a different number of objects.

Techniques for generalized planning \cite{Winner07loopdistill:learning, DBLP:conf/aaai/SrivastavaIZ08, DBLP:conf/aips/BonetPG09, DBLP:journals/ai/SrivastavaIZ11} primarily focus on computing algorithm-like plans
that can be used to solve a broad class of problems. These approaches require action models and do not generate heuristics, instead
the plan itself is executed for an arbitrary number of objects.

Our approach for synthesizing domain-independent HGFs differs from these prior efforts along multiple dimensions. Instead of relying on specialized network blocks, we use a rich input representation that is model-agnostic i.e. independent of action models. Using canonical abstractions, we abstract away problem
dependent information like object names but retain the ability to capture the state structure. In the absence of training data, leapfrogging can be used to incrementally generate new training data and gradually improve the quality of the learned heuristic.

\section{Conclusions}
\label{sec:conclusion}
We presented a new approach for learning heuristic generating functions when access to symbolic action models are unavailable.  GHNs demonstrate the
capability to generalize to a very large number of objects, initial states and goal conditions. In the
absence of training data, leapfrogging can be used to incrementally learn better heuristics. Our results conclude that this approach is promising for future investigation.
There is much to explore in canonical abstraction and deep neural networks for learning heuristics and we
believe that our results might motivate future research.
\section{Ethics Statement}

Automated planning is widely regarded as one of the longstanding problems of AI.
This research would enable autonomous agents to carry out automated planning in the absence of domain experts. We believe that this would improve the accessibility of AI systems, as it would would allow non-expert users to assign AI systems new tasks efficiently without having to invest in an AI expert who could create a symbolic domain representation and also a heuristic generating function.

Our approach for planning comes with guarantees of soundness and completeness. This implies that it will find a solution if there exists one, and the solution that it finds will be correct as per the simulator's action encodings. As in any approach that uses simulators, this method is susceptible to errors in programming and in simulator design. This can be addressed independently through research on formal verification of simulators used in AI.

\bibliography{references}

\begin{thebibliography}{36}
\providecommand{\natexlab}[1]{#1}
\providecommand{\url}[1]{\texttt{#1}}
\providecommand{\urlprefix}{URL }
\expandafter\ifx\csname urlstyle\endcsname\relax
  \providecommand{\doi}[1]{doi:\discretionary{}{}{}#1}\else
  \providecommand{\doi}{doi:\discretionary{}{}{}\begingroup
  \urlstyle{rm}\Url}\fi

\bibitem[{Alkhazraji et~al.(2020)Alkhazraji, Frorath, Gr{\"u}tzner, Helmert,
  Liebetraut, Mattm{\"u}ller, Ortlieb, Seipp, Springenberg, Stahl, and
  W{\"u}lfing}]{alkhazraji-et-al-zenodo2020}
Alkhazraji, Y.; Frorath, M.; Gr{\"u}tzner, M.; Helmert, M.; Liebetraut, T.;
  Mattm{\"u}ller, R.; Ortlieb, M.; Seipp, J.; Springenberg, T.; Stahl, P.; and
  W{\"u}lfing, J. 2020.
\newblock Pyperplan.

\bibitem[{Arfaee, Zilles, and Holte(2010)}]{DBLP:conf/socs/ArfaeeZH10}
Arfaee, S.~J.; Zilles, S.; and Holte, R.~C. 2010.
\newblock Bootstrap Learning of Heuristic Functions.
\newblock In \emph{Proceedings of the 3rd Annual Symposium on Combinatorial
  Search, {SOCS}}.

\bibitem[{Bengio et~al.(2009)Bengio, Louradour, Collobert, and
  Weston}]{DBLP:conf/icml/BengioLCW09}
Bengio, Y.; Louradour, J.; Collobert, R.; and Weston, J. 2009.
\newblock Curriculum learning.
\newblock In \emph{Proceedings of the 26th Annual International Conference on
  Machine Learning, {ICML}}.

\bibitem[{Bonet, Franc{\`{e}}s, and Geffner(2019)}]{DBLP:conf/aaai/BonetFG19}
Bonet, B.; Franc{\`{e}}s, G.; and Geffner, H. 2019.
\newblock Learning Features and Abstract Actions for Computing Generalized
  Plans.
\newblock In \emph{Proceedings of the 33rd {AAAI} Conference on Artificial
  Intelligence, {AAAI}}.

\bibitem[{Bonet and Geffner(2001)}]{DBLP:journals/ai/BonetG01}
Bonet, B.; and Geffner, H. 2001.
\newblock Planning as heuristic search.
\newblock \emph{Artif. Intell.} 129(1-2): 5--33.

\bibitem[{Bonet, Palacios, and Geffner(2009)}]{DBLP:conf/aips/BonetPG09}
Bonet, B.; Palacios, H.; and Geffner, H. 2009.
\newblock Automatic Derivation of Memoryless Policies and Finite-State
  Controllers Using Classical Planners.
\newblock In \emph{Proceedings of the 19th International Conference on
  Automated Planning and Scheduling, {ICAPS}}.

\bibitem[{Bylander(1991)}]{DBLP:conf/ijcai/Bylander91}
Bylander, T. 1991.
\newblock Complexity Results for Planning.
\newblock In \emph{Proceedings of the 12th International Joint Conference on
  Artificial Intelligence, {IJCAI}}.

\bibitem[{Bylander(1994)}]{DBLP:journals/ai/Bylander94}
Bylander, T. 1994.
\newblock The Computational Complexity of Propositional {STRIPS} Planning.
\newblock \emph{Artif. Intell.} 69(1-2): 165--204.

\bibitem[{Celorrio, Aguas, and Jonsson(2019)}]{DBLP:journals/ker/CelorrioAJ19}
Celorrio, S.~J.; Aguas, J.~S.; and Jonsson, A. 2019.
\newblock A review of generalized planning.
\newblock \emph{Knowledge Eng. Review} 34: e5.

\bibitem[{Celorrio et~al.(2012)Celorrio, de~la Rosa, Fern{\'{a}}ndez,
  Fern{\'{a}}ndez{-}Rebollo, and Borrajo}]{DBLP:journals/ker/JimenezRFFB12}
Celorrio, S.~J.; de~la Rosa, T.; Fern{\'{a}}ndez, S.;
  Fern{\'{a}}ndez{-}Rebollo, F.; and Borrajo, D. 2012.
\newblock A review of machine learning for automated planning.
\newblock \emph{Knowledge Eng. Review} 27(4): 433--467.

\bibitem[{Chollet et~al.(2015)}]{chollet2015keras}
Chollet, F.; et~al. 2015.
\newblock Keras.
\newblock \url{https://keras.io}.

\bibitem[{Fawcett et~al.(2011)Fawcett, Helmert, Hoos, Karpas, R{\"o}ger, and
  Seipp}]{source:generator}
Fawcett, C.; Helmert, M.; Hoos, H.; Karpas, E.; R{\"o}ger, G.; and Seipp, J.
  2011.
\newblock {FD-Autotune}: Domain-Specific Configuration using {Fast} {Downward}.

\bibitem[{Fikes and Nilsson(1971)}]{DBLP:conf/ijcai/FikesN71}
Fikes, R.; and Nilsson, N.~J. 1971.
\newblock {STRIPS:} {A} New Approach to the Application of Theorem Proving to
  Problem Solving.
\newblock In \emph{Proceedings of the 2nd International Joint Conference on
  Artificial Intelligence, {IJCAI}}.

\bibitem[{Fox and Long(2003)}]{DBLP:journals/jair/FoxL03}
Fox, M.; and Long, D. 2003.
\newblock {PDDL2.1:} {A}n Extension to {PDDL} for Expressing Temporal Planning
  Domains.
\newblock \emph{J. Artif. Intell. Res.} 20: 61--124.

\bibitem[{Groshev et~al.(2018{\natexlab{a}})Groshev, Goldstein, Tamar,
  Srivastava, and Abbeel}]{DBLP:conf/aips/GroshevGTSA18}
Groshev, E.; Goldstein, M.; Tamar, A.; Srivastava, S.; and Abbeel, P.
  2018{\natexlab{a}}.
\newblock Learning Generalized Reactive Policies Using Deep Neural Networks.
\newblock In \emph{Proceedings of the 28th International Conference on
  Automated Planning and Scheduling, {ICAPS}}.

\bibitem[{Groshev et~al.(2018{\natexlab{b}})Groshev, Tamar, Goldstein,
  Srivastava, and Abbeel}]{DBLP:conf/aaaiss/GroshevTGSA18}
Groshev, E.; Tamar, A.; Goldstein, M.; Srivastava, S.; and Abbeel, P.
  2018{\natexlab{b}}.
\newblock Learning Generalized Reactive Policies using Deep Neural Networks.
\newblock In \emph{Proceedings of the 28th International Conference on
  Automated Planning and Scheduling, {ICAPS}}.

\bibitem[{Hart, Nilsson, and Raphael(1968)}]{DBLP:journals/tssc/HartNR68}
Hart, P.~E.; Nilsson, N.~J.; and Raphael, B. 1968.
\newblock A Formal Basis for the Heuristic Determination of Minimum Cost Paths.
\newblock \emph{{IEEE} Trans. Systems Science and Cybernetics} 4(2): 100--107.

\bibitem[{Helmert(2006)}]{DBLP:journals/jair/Helmert06}
Helmert, M. 2006.
\newblock The Fast Downward Planning System.
\newblock \emph{J. Artif. Intell. Res.} 26: 191--246.

\bibitem[{Helmert and Domshlak(2009)}]{DBLP:conf/aips/HelmertD09}
Helmert, M.; and Domshlak, C. 2009.
\newblock Landmarks, Critical Paths and Abstractions: What's the Difference
  Anyway?
\newblock In \emph{Proceedings of the 19th International Conference on
  Automated Planning and Scheduling, {ICAPS}}.

\bibitem[{Hinton, Srivastava, and Swersky(2012)}]{RMSProp}
Hinton, G.; Srivastava, N.; and Swersky, K. 2012.
\newblock RMSProp.
\newblock
  \url{https://www.cs.toronto.edu/~tijmen/csc321/slides/lecture_slides_lec6.pdf}.

\bibitem[{Hoffmann and Nebel(2001)}]{DBLP:journals/jair/HoffmannN01}
Hoffmann, J.; and Nebel, B. 2001.
\newblock The {FF} Planning System: Fast Plan Generation Through Heuristic
  Search.
\newblock \emph{J. Artif. Intell. Res.} 14: 253--302.

\bibitem[{Long and Fox(2003)}]{DBLP:journals/jair/LongF03}
Long, D.; and Fox, M. 2003.
\newblock The 3rd International Planning Competition: Results and Analysis.
\newblock \emph{J. Artif. Intell. Res.} 20: 1--59.

\bibitem[{Richter and Westphal(2010)}]{DBLP:journals/jair/RichterW10}
Richter, S.; and Westphal, M. 2010.
\newblock The {LAMA} Planner: Guiding Cost-Based Anytime Planning with
  Landmarks.
\newblock \emph{J. Artif. Intell. Res.} 39: 127--177.

\bibitem[{Rivlin, Hazan, and Karpas(2020)}]{arxiv:GPDRL}
Rivlin, O.; Hazan, T.; and Karpas, E. 2020.
\newblock Generalized Planning With Deep Reinforcement Learning.
\newblock \emph{CoRR} abs/2005.02305.

\bibitem[{Russell and Norvig(2010)}]{DBLP:books/daglib/0023820}
Russell, S.~J.; and Norvig, P. 2010.
\newblock \emph{Artificial Intelligence - {A} Modern Approach, Third
  International Edition}.
\newblock Pearson Education.

\bibitem[{Sagiv, Reps, and Wilhelm(2002)}]{DBLP:journals/toplas/SagivRW02}
Sagiv, S.; Reps, T.~W.; and Wilhelm, R. 2002.
\newblock Parametric shape analysis via 3-valued logic.
\newblock \emph{{ACM} Trans. Program. Lang. Syst.} 24(3): 217--298.

\bibitem[{Sanner(2010)}]{Sanner:RDDL}
Sanner, S. 2010.
\newblock Relational Dynamic Influence Diagram Language ({RDDL}): Language
  Description.
\newblock \url{http://users.cecs.anu.edu.au/~ssanner/IPPC_2011/RDDL.pdf}.

\bibitem[{Shen, Trevizan, and Thi{\'e}baux(2020)}]{shen20:stripshgn}
Shen, W.; Trevizan, F.; and Thi{\'e}baux, S. 2020.
\newblock Learning Domain-Independent Planning Heuristics with Hypergraph
  Networks.
\newblock In \emph{Proceedings of the 30th International Conference on
  Automated Planning and Scheduling, {ICAPS}}.

\bibitem[{Srivastava, Immerman, and
  Zilberstein(2008)}]{DBLP:conf/aaai/SrivastavaIZ08}
Srivastava, S.; Immerman, N.; and Zilberstein, S. 2008.
\newblock Learning Generalized Plans Using Abstract Counting.
\newblock In \emph{Proceedings of the 23rd {AAAI} Conference on Artificial
  Intelligence, {AAAI}}.

\bibitem[{Srivastava, Immerman, and
  Zilberstein(2011)}]{DBLP:journals/ai/SrivastavaIZ11}
Srivastava, S.; Immerman, N.; and Zilberstein, S. 2011.
\newblock A new representation and associated algorithms for generalized
  planning.
\newblock \emph{Artif. Intell.} 175(2): 615--647.

\bibitem[{Srivastava et~al.(2014)Srivastava, Russell, Ruan, and
  Cheng}]{DBLP:conf/uai/SrivastavaRRC14}
Srivastava, S.; Russell, S.~J.; Ruan, P.; and Cheng, X. 2014.
\newblock First-Order Open-Universe POMDPs.
\newblock In \emph{Proceedings of the 30th Conference on Uncertainty in
  Artificial Intelligence, {UAI}}.

\bibitem[{Tamar et~al.(2016)Tamar, Levine, Abbeel, Wu, and
  Thomas}]{DBLP:conf/nips/TamarLAWT16}
Tamar, A.; Levine, S.; Abbeel, P.; Wu, Y.; and Thomas, G. 2016.
\newblock Value Iteration Networks.
\newblock In \emph{Annual Conference on Neural Information Processing Systems}.

\bibitem[{Toyer et~al.(2018)Toyer, Trevizan, Thi{\'{e}}baux, and
  Xie}]{DBLP:conf/aaai/ToyerTTX18}
Toyer, S.; Trevizan, F.~W.; Thi{\'{e}}baux, S.; and Xie, L. 2018.
\newblock Action Schema Networks: Generalised Policies With Deep Learning.
\newblock In \emph{Proceedings of the 32nd {AAAI} Conference on Artificial
  Intelligence, {AAAI}}.

\bibitem[{Winner and Veloso(2003)}]{DBLP:conf/icml/WinnerV03}
Winner, E.; and Veloso, M.~M. 2003.
\newblock {DISTILL:} Learning Domain-Specific Planners by Example.
\newblock In \emph{Proceedings of the 20th International Conference on Machine
  Learning, {ICML}}.

\bibitem[{Winner and Veloso(2007)}]{Winner07loopdistill:learning}
Winner, E.~Z.; and Veloso, M. 2007.
\newblock Loopdistill: Learning domain-specific planners from example plans.
\newblock In \emph{ICAPS Workshop on Planning and Scheduling}.

\bibitem[{Yoon, Fern, and Givan(2007)}]{DBLP:conf/ijcai/YoonFG07}
Yoon, S.~W.; Fern, A.; and Givan, R. 2007.
\newblock Using Learned Policies in Heuristic-Search Planning.
\newblock In \emph{Proceedings of the 20th International Joint Conference on
  Artificial Intelligence, {IJCAI}}.

\end{thebibliography}

\end{document}